\documentclass[a4paper,twoside,final]{article}

\usepackage{epsfig}
\usepackage{subcaption}
\usepackage{calc}
\usepackage{amssymb}
\usepackage{amstext}
\usepackage{amsmath}
\usepackage{amsthm}
\usepackage{multicol}
\usepackage{pslatex}
\usepackage{apalike}
\usepackage{algorithm2e}
\usepackage{url}
\usepackage[bottom]{footmisc}
\usepackage{multirow}
\usepackage{geometry}
\usepackage{tabularx}
\usepackage{booktabs}
\usepackage{array}
\usepackage{caption}
\usepackage{ragged2e}
\usepackage{threeparttable}

\usepackage{hyperref}
\usepackage{dcolumn}
\usepackage{tabularx,booktabs}
\usepackage{SCITEPRESS}     

\def\valrange{[0,1]}
\def\imgrange{{C \times H \times W}}
\def\vidrange{{N \times C \times H \times W}}
\def\methodname{{IC2VQA}}

\begin{document}

\title{Cross-Modal Transferable Image-to-Video Attack on Video Quality Metrics}

\author{\authorname{Georgii Gotin\sup{1}\orcidAuthor{0009-0007-7176-703X}, Ekaterina Shumitskaya\sup{2,3,1}\orcidAuthor{0000-0002-6453-5616}, Anastasia Antsiferova\sup{3,2,4}\orcidAuthor{0000-0002-1272-5135} \\and Dmitriy Vatolin\sup{1,2,3}\orcidAuthor{0000-0002-8893-9340}}
\affiliation{\sup{1}Lomonosov Moscow State University, Moscow, Russia}
\affiliation{\sup{2}ISP RAS Research Center for Trusted Artificial Intelligence, Moscow, Russia}
\affiliation{\sup{3}MSU Institute for Artificial Intelligence, Moscow, Russia}
\affiliation{\sup{4}Laboratory of Innovative Technologies for Processing Video Content, Innopolis University, Innopolis, Russia}
\email{georgii.gotin@graphics.cs.msu.ru}
}

\keywords{video quality assessment, video quality metric, adversarial attack, cross-modal, CLIP}

\abstract{Recent studies have revealed that modern image and video quality assessment (IQA/VQA) metrics are vulnerable to adversarial attacks. An attacker can manipulate a video through preprocessing to artificially increase its quality score according to a certain metric, despite no actual improvement in visual quality. Most of the attacks studied in the literature are white-box attacks, while black-box attacks in the context of VQA have received less attention. Moreover, some research indicates a lack of transferability of adversarial examples generated for one model to another when applied to VQA. In this paper, we propose a cross-modal attack method, \methodname, aimed at exploring the vulnerabilities of modern VQA models. This approach is motivated by the observation that the low-level feature spaces of images and videos are similar. We investigate the transferability of adversarial perturbations across different modalities; specifically, we analyze how adversarial perturbations generated on a white-box IQA model with an additional CLIP module can effectively target a VQA model. The addition of the CLIP module serves as a valuable aid in increasing transferability, as the CLIP model is known for its effective capture of low-level semantics. Extensive experiments demonstrate that \methodname achieves a high success rate in attacking three black-box VQA models. We compare our method with existing black-box attack strategies, highlighting its superiority in terms of attack success within the same number of iterations and levels of attack strength. We believe that the proposed method will contribute to the deeper analysis of robust VQA metrics.
}

\onecolumn \maketitle \normalsize \setcounter{footnote}{0} \vfill

\begin{figure*}[!h]
  \centering
   {\includegraphics[width=\linewidth]{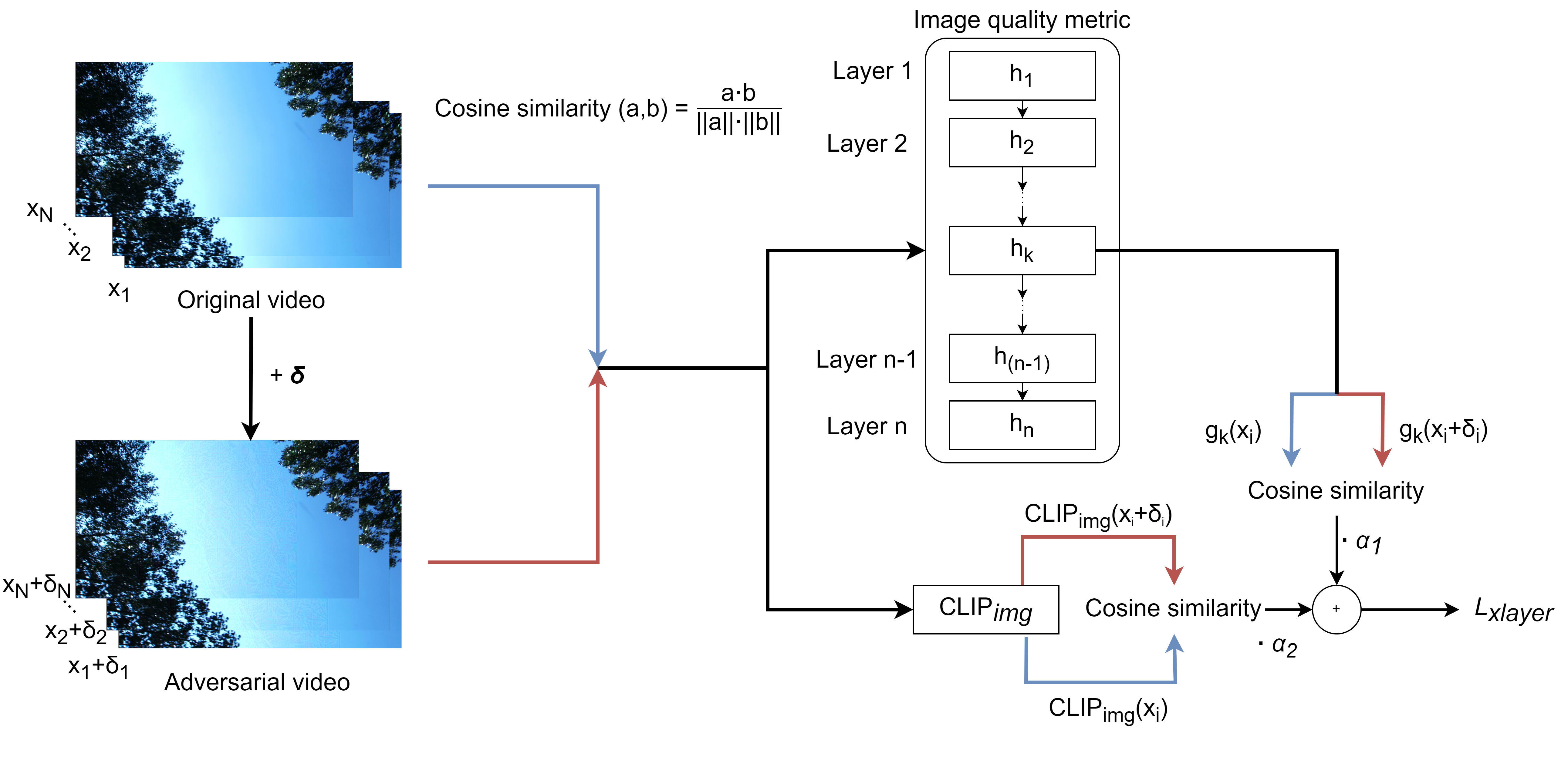}}
  \caption{Scheme of the proposed {\methodname} method. Given an original video, each clip runs through image quality metric with saving of output on the k-th layer and through CLIP image model with saving full output. After that attacked video runs same models with saving same outputs. Then cosine similarities of saved outputs are respectively aggregated in cross layer loss.}
  \label{fig:diagram1}
 \end{figure*}

\section{\uppercase{Introduction}}
\label{sec:introduction}

Modern No-Reference Video Quality Assessment (NR-VQA) metrics are vulnerable to adversarial attacks \cite{yang2024exploring},  \cite{yang2024beyond}, \cite{zhang2024vulnerabilities}, \cite{siniukov2023unveiling}, \cite{SHUMITSKAYA2024103913}. This raises concerns about the safety of relying on these metrics to automatically assess video quality in real-world scenarios, such as public benchmarks and in more critical situations, such as autonomous driving. Adversarial attacks on VQA metrics can be classified into two categories: white-box and black-box attacks. White-box attacks operate with complete access to the VQA metric, including its architecture and gradients. In contrast, black-box attacks works without any knowledge of the metric's architecture and can only send queries to receive the metric's response. There is also a subclass of black-box attacks that utilizes a proxy white-box model to generate adversarial perturbations. These generated perturbations can effectively deceive unseen models in black-box settings. However, in \cite{zhang2022perceptual} the authors demonstrated that VQA metrics exhibit poor transferability across different models. This limitation may appear from the fact that VQA models place significant emphasis on various texture and noise details, which can vary greatly among different models. In contrast, classification tasks typically focus primarily on the semantic content of images, leading to greater consistency in performance across diverse classification models. In other words, creating a transferable attack for VQA metrics is more challenging than for classification tasks. To address this issue, we propose transferable cross-model attack to perform white-box attack on \textbf{I}mage quality metric and \textbf{C}LIP and transfer it to \textbf{V}ideo \textbf{Q}uality \textbf{A}ssessment model \textbf{(IC2VQA)}.
Figure \ref{fig:diagram1} provides an overview of the proposed \methodname 
 method. {\methodname} takes individual frames of the original video and generates adversarial noise for each frame.

 Our main contributions are as follows.
\begin{itemize}
\item We propose a novel method for transferable cross-modal attacks on NR VQA metrics that utilizes IQA metrics and CLIP model
\item We conduct comprehensive experiments using 12 high-resolution videos and 3 target VQA models and show the superiority of the proposed method among existing methods
\item We analyze the correlations between features in the deep layers of IQA and VQA metrics 
\item We made our code available on GitHub:\\ \url{https://github.com/GeorgeGotin/IC2VQA}.
\end{itemize}

\section{\uppercase{Related work}}

\subsection{Image- and Video-Quality Metrics}
Image and video quality assessment (IQA/VQA) metrics can be divided into full-reference and no-reference (also known as blind IQA/VQA). Full-reference quality metrics compare two images/videos, while no-reference metrics assess the visual quality of a single image/video. These tasks are fundamentally different: full-reference IQA focuses on measuring distances between images in various feature spaces, while no-reference IQA evaluates the quality of an image based solely on the distorted image. No-reference image- and video-quality assessment (NR-VQA) metrics fall into distortion-specific and general-purpose. Distortion-specific approaches predict the quality score for a particular type of distortion, such as compression \cite{wang2015no} or blurring \cite{chen2011no}. However, these methods have limited real-world applications because it is not always possible to specify the type of distortion. They may not capture the complex mixtures of distortions that often occur in real-world images and videos. However, general-purpose NR-VQA approaches assess the image quality of any distortion. In this work, we focus on the problem of attacking NR-VQA metrics \cite{li2019quality}, \cite{li2021unified}, \cite{zhang2022texture} to find metrics that are robust to transferable cross-modal attacks.

\subsection{Adversarial Attacks on Image- and Video-Quality Metrics}

The problem of vulnerability analysis of novel NR IQA models to adversarial attacks was widely discussed in previous works: \cite{yang2024exploring}, \cite{leonenkova2024ti}, \cite{kashkarov2024can}, \cite{deng2024sparse}, \cite{konstantinov2024image}, \cite{yang2024beyond}, \cite{zhang2024vulnerabilities}, \cite{ran2025black}, \cite{meftah2023evaluating}, \cite{siniukov2023unveiling}, \cite{pmlr-v235-shumitskaya24a}, \cite{SHUMITSKAYA2024103913}.
Some works have been conducted as part of the MediaEval task: ``Pixel Privacy: Quality Camouflage for Social Images'' \cite{mediaeval}, where participants aimed to improve image quality while reducing the predicted quality score. This task is similar to the vanilla adversarial attack on quality metrics, but to decrease the score rather than increase it. In \cite{bonnet2020fooling}, the authors generated adversarial examples for NR models using PGD attack \cite{madry2018towards}. Zhao et al. \cite{zhao2023adversarial} proposed to attack NR metrics by applying image transformations based on optimizing a human-perceptible color filter. They also demonstrated that this attack is even resistant to JPEG compression. However, these studies are limited to small-scale experiments and lack in-depth analysis. Several comprehensive works have recently been published that systematically investigate adversarial attacks against NR models. 

In \cite{zhang2022perceptual}, a two-step perceptual attack was introduced for the NR metrics. The authors established the attack's goal as a Lagrangian function that utilizes some FR metric, which acts as a ``perceptual constraint'', alongside the NR metric representing the target model. By adjusting the Lagrange multiplier, they produced a range of perturbed images that exhibited varying degrees of visibility regarding their distortions.
Their extensive experiments demonstrated that the proposed attack effectively deceived four different NR metrics; however, the adversarial examples did not transfer well across various models, indicating specific design vulnerabilities within the NR metrics assessed. In \cite{Shumitskaya_2022_BMVC}, the authors trained the UAP on low-resolution data and then applied it to high-resolution data. This method significantly reduces the time required to attack videos, as it requires only adding perturbations to individual frames. In the study by \cite{korhonen2022adversarial}, the authors create adversarial perturbations for NR metrics by injecting the perturbations into textured areas using the Sobel filter. They also demonstrated that adversarial images generated for a simple NR metric in white-box settings are transferable and can deceive several NR metrics with more complex architecture in black-box settings. In \cite{antsiferova2024comparing}, the authors presented a methodology for evaluating the robustness of NR and FR IQA metrics through a wide range of adversarial attacks and released an open benchmark.

To the best of our knowledge, no methods have been designed for transferable cross-modal attacks from NR IQA to NR VQA metrics, which is a subject of this work.

\subsection{Transferable Attacks on Image Classification}
Adversarial attacks have received significant attention in the domain of machine learning, particularly in image classification tasks. The phenomenon of transferability, where adversarial examples generated on one model can deceive another (potentially different) model, has been investigated in many works. Papernot et al. 
 \cite{papernot2016transferability} explored this aspect and demonstrated that transferability is a useful property that could be exploited in black-box settings, where the attacker has limited knowledge of the target model. They also experimentally showed that adversarial examples could be trained on weaker models and successfully deceive more robust classifiers. Various methods have been proposed to enhance the effectiveness of transferable attacks. Some of them \cite{xie2019improving}, \cite{lin2019nesterov}, \cite{dong2019evading} apply data-augmentation techniques to enhance the generalization of adversarial examples and reduce the risk of overfitting the white-box model. For example, the translation-invariant attack \cite{dong2019evading} executes slight horizontal and vertical shifts of the input. The second direction to improve transferability is to modify the gradients used to update adversarial perturbations \cite{dong2019evading}, \cite{lin2019nesterov}, \cite{wu2020skip}. For example, the momentum iterative attack \cite{dong2019evading} stabilizes the update directions using the addition of momentum in the iterative process. The third approach concentrates on disrupting the shared classification properties among different models \cite{wu2020boosting}, \cite{huang2019enhancing}, \cite{lu2020enhancing}. One example is the Attention-guided attack \cite{wu2020boosting}, which prioritizes the corruption of critical features that are commonly utilized by various architectures. Recently, innovative cross-modal approaches have been proposed that leverage the correlations between spatial features encoded by different modalities \cite{wei2022cross}, \cite{chen2023gcma}, \cite{yang2025prompt}. Image2Video attack, proposed in \cite{wei2022cross}, is an attack to successfully transfer from image to video recognition models. 

\section{\uppercase{Proposed Method}}

\subsection{Problem Formulation}
Let’s consider we have a video $x \in X \subset \valrange^\vidrange$ , where $N$ --- number of frames in video, $C$ --- number of channels in video, $H$, $W$ --- height and width of video respectively, $X$ is the set of all possible videos. We define video quality metric as $f: X \rightarrow [0;1]$, image quality metric as $g: \valrange^\imgrange \xrightarrow{} [0;1]$. Image quality metric can be expressed in layered form as $g = h_K \circ h_{K-1}\circ...\circ h_1$, so 
\begin{equation}
\label{layered_image_metric}
    g(x_i) = h_K(h_{K-1}(...h_1(x_i)...)),
\end{equation}
where each function $h_k:P_{k-1} \rightarrow P_k$ corresponds to a processing layer with $P_0=\valrange^\imgrange$ being the input feature space and $P_K=[0,1]$ being the output range of the metric. $g_k$ defines the composition of the first $k$ layers:
\begin{equation}
\label{g_k}
\begin{split}
    g_k=h_k\circ...\circ h_1 \\
    g_k:\valrange^\imgrange\rightarrow P_k.
\end{split}
\end{equation}
Each $g_k$ serves the $k$-th layer of the quality metric, where $P_k$ represents the feature spaces corresponding to that layer.

\subsection{Method}

The primary goal of the attack is to make the predicted quality score of video $f(x+\delta)$ on the attacked video deviate from the original score $f(x)$, where $\delta$ is the perturbation on the video x. Also, the rank of correlation of predicted score with MOS is important, so our goal is to shrink it as possible. This method was based on method proposed by Zhipeng Wei as I2V\cite{wei2022cross}. 

The proposed attack is designed to mislead the video quality metric. It creates adversarial frame $\delta_i \in \valrange^\imgrange$ for each $i$-th frame on the input video. 
To maintain the imperceptible of this adversarial perturbation, we import a constraint on its magnitude 
$\|\delta\|_p\le\varepsilon$, where 
$\|\cdot\|_p$ denotes $L_p$ norm. In our research, we adopt the $L_\infty$ norm due to its computational efficiency compared to other $L_p$ norms.

Based on observation of correlations between layers of video and image quality metrics, we proposed the cross-layer loss, this loss is designed to influence the features of the layers within the image quality metric and enhance it's the effectiveness in black box settings.
The cross-layer loss of the $k$-th layer defined as follows 
\begin{equation}
    \label{loss_xlayer}
    \mathcal{L}_{xlayer}=\frac{1}{N}\sum_{i=1}^{N}\frac{g_k(x_i+\delta_i) \cdot g_k(x_i)}{\|g_k(x_i+\delta_i)\| \|g_k(x_i)\|},
\end{equation}
where $x$ --- the original video with $N$ frames, $x_i$ --- $i$-th frame of the video. 
We propose multi-modal cross-layer loss for better implementation and generalization across different feature domains. This loss utilizes adversarial perturbation $\delta$ to simultaneously optimize an ensemble of image quality metrics $g^{(1)},..,g^{(F)}$ with layers $k_f$ for $f$-th metric. Consequently, the overall cross-layer loss can be defined as follows: 
\begin{equation}
\begin{array}{cc}
     \mathcal{L}_{sim} = \frac{1}{N}\sum\limits_{i=1}^{N}\sum\limits_{f=1}^{F}\alpha_{f}\frac{g^{(f)}_k(x_i+\delta_i) \cdot g^{(f)}_k(x_i)}{\|g^{(f)}_k(x_i+\delta_i)\|\|g^{(f)}_k(x_i)\|} + \\ + 
    \frac{1}{F}\sum\limits_{f=1}^{F}\|1-\alpha_f\|,
\end{array}
\label{loss_xlayer_sim}
\end{equation}
where $\alpha_f$ --- constant positive value, initialized with ones.

\begin{figure}[!h]
  \centering
   {\includegraphics[width=\linewidth]{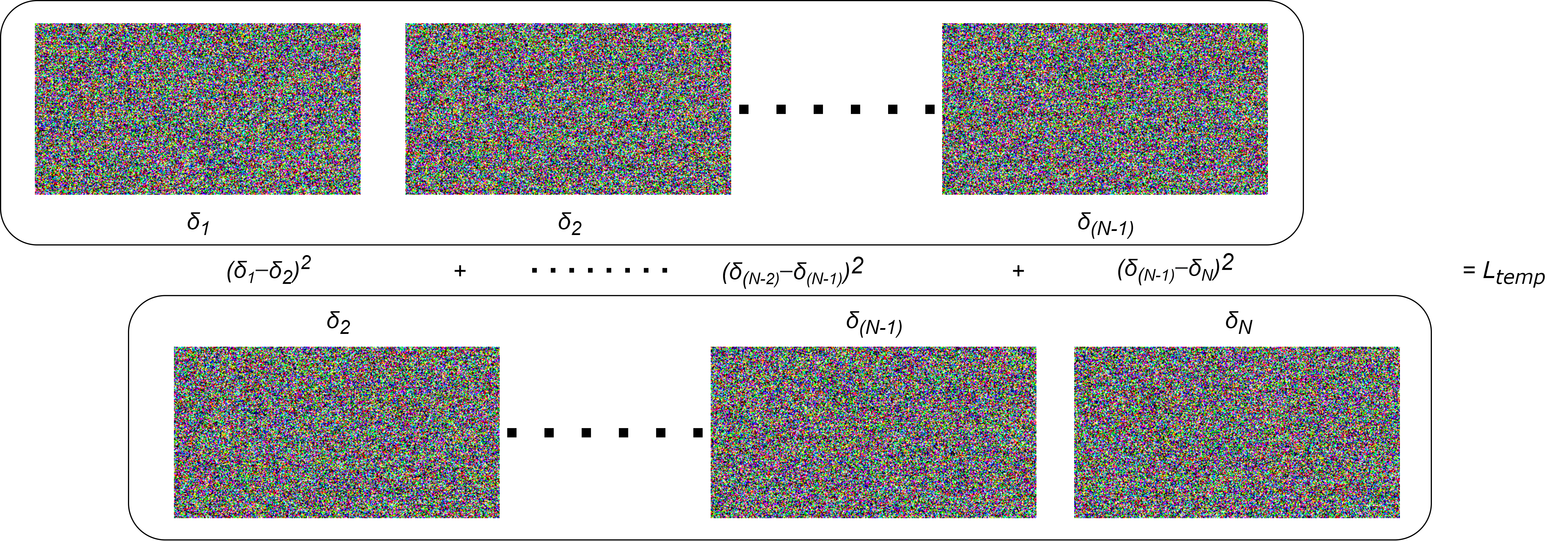}}
  \caption{Overview of the temporal loss computing. For each pair of frames from original and attacked videos difference $\Delta$ is computed. The temporal loss is computed as square root of sum of all differences.}
  \label{fig:diagram2}
 \end{figure}

To enhance temporal stability of the attacked video $x + \delta$ and further ensure that the adversarial perturbation $\delta$ is imperceptible, we added a temporal loss component (Figure \ref{fig:diagram2})
\begin{equation}
    \label{loss_temp}
    \mathcal{L}_{temp}=\frac{1}{N-1}\sum_{i=1}^{N-1}\|\delta_{i+1}-\delta_i\|_2.
\end{equation}


\subsection{Algorithm}

\begin{algorithm}[!h]
 \caption{Algorithm of the consistent attack with multiple image quality metrics}
 \label{alg_cons}
 \KwData{original video $x \in \valrange^\vidrange$, F image quality metrics $g_1,...,g_F:\valrange^\imgrange \rightarrow[0,1]$, $k_1,...,k_F$ --- number of the layer, perturbation budget $\varepsilon$, number of iterations $I$}
 \KwResult{$\delta \in \valrange^\vidrange, s.t. \|\delta\|\le\varepsilon$}
 $\delta = (1/255)^\vidrange$\;
  \For{i from 1 to I}{
    \For{f from 1 to F}{
        Calculate $\mathcal{L}_{xlayer}$ as\ in\ \ref{loss_xlayer} for $g^{(f)}$\;
        
        Calculate $\mathcal{L}_{temp}$ as\ in\ \ref{loss_temp}\;
        
        $\delta \leftarrow ADAM(\alpha, loss_{xlayer} + loss_{temp})$;
        
        $\delta \leftarrow clip_{\varepsilon}(\delta)$;
    }
  }
\end{algorithm}

We construct our attack as presented in Algorithm \ref{alg_cons}, which is applied to image quality metrics. At each step of the attack, the cross-layer loss for the $f$-th image quality metric is computed and the adversarial noise is optimized using the Adam optimizer. Subsequently, the noise is clipped to ensure it remains within the bounds of $\varepsilon$ according to the $L_{\infty}$ norm. Experiments shown that alternative version of algorithms, where all losses are summed with weights as described in (\ref{loss_xlayer_sim}), yields lower scores compared to the final algorithm. By applying this algorithm to attacks targeting a single image quality metric, IC2VQA has effectively transformed into a single-metric attack.

\section{\uppercase{Experiments}}
\subsection{Dataset}
We evaluate our attack using a subset of Xiph.org (Derf’s) dataset \cite{xiph}. The subset contains ten videos downscaled from 1080p to 540p and trimmed to 75 frames. The videos have different patterns of image and motions in it such as shooting from a tripod, moving crowd, running water, etc. 

\subsection{Quality Metrics}

\subsubsection{Image Quality Metrics}
For ensembles of image quality metrics we used NIMA \cite{talebi2018nima}, PaQ-2-PiQ \cite{ying2020patches}, SPAQ \cite{fang2020perceptual} metrics. To further boost the transferability we added additional modalities such as CLIP model \cite{radford2021learning}. To get feature vectors, in NIMA model attack utilizes layers after classifier and after global pool, in PaQ-2-PiQ model attack utilizes layers after roi-pool layer and body, in SPAQ model attack utilizes first, second, third and fourth layers. In CLIP model, an output of the CLIP image module was utilized.

\subsubsection{Video Quality Metrics}
As black-boxed video-metric we used the VSFA \cite{li2019quality}, MDTVSFA \cite{li2021unified} and TiVQA \cite{zhang2022texture} trained on the KoNViD-1k \cite{hosu2020konstanz}. These metrics evaluate quality scores by taking into account both the spatial and temporal characteristics of the videos. 

\subsection{Comparison with Other Methods}
Due to the lack of existing black-box image-to-video quality model attacks, we compared our method against one transferable attack, the PGD attack \cite{madry2018towards}, adapted for image-to-video scenarios, as well as two black-box attacks: Square Attack \cite{andriushchenko2020square} and AttackVQA \cite{zhang2024vulnerabilities}. The latter was specifically designed to target VQA metrics. For comparison, we tested all methods using a grid of parameters for $\epsilon$ and $I$ to generate attacked videos with varying levels of distortion. Recall that $\epsilon$ represents the $L_\infty$ norm restriction on generated perturbation and $I$ is the number of iterations used for attack. Next, we measured the VQA metric scores of the attacked videos and calculated the correlations between these scores and a corresponding linearly decreasing vector. As the $\epsilon$/$I$ parameters increase while keeping $I$/$\epsilon$ fixed, the quality of the attacked videos tends to degrade in an approximately linear manner. Therefore, an effective VQA metric should exhibit a strong correlation with this vector for the attacked videos. Consequently, if the metric is vulnerable, it will be indicated by a low correlation. Additionally, the most effective attacks will result in lower correlations, so we assess attack success by evaluating their ability to reduce these correlations. In our experiments, we used absolute values Pearson's (PLCC) and Spearman's (SRCC)  correlations.


\subsection{Parameters}
We evaluated the proposed and comparison methods using a range of $\epsilon$ and $I$ parameters to assess their effectiveness under various conditions. We used the following grids of parameters: $\epsilon$~=~[1/255,2/255,5/255,10/255,15/255,20/255,50/255] and $I$~=~[1,2,5,10,20].

\section{\uppercase{Results}}

\begin{table*}[h]
\caption{Comparison of the proposed transferable cross-modal \methodname attack with two black-box attacks (Square Attack \cite{andriushchenko2020square} and AttackVQA \cite{zhang2024vulnerabilities}) and one transferable PGD attack \cite{madry2018towards} targeting three VQA metrics. The table presents the mean absolute values of PLCC and SROCC correlations across different epsilons between linearly decreasing vectors and attacked VQA scores. For each score  Transferable attacks were performed using three different white-box IQA metrics. }
\label{table comparison}
\begin{center}
\begin{tabular}{ccccc}
\toprule
\multicolumn{1}{c}{Attack} & \multicolumn{1}{c}{Image quality metric$^\ast$} & \multicolumn{3}{c}{Video metric} \\
\cline{3-5}
\multicolumn{2}{c}{} & \multicolumn{1}{c}{VSFA} & \multicolumn{1}{c}{MDTVSFA} & \multicolumn{1}{c}{TiVQA}\\
\multicolumn{2}{c}{} & \multicolumn{1}{c}{PLCC$\downarrow$ / SRCC$\downarrow$} & \multicolumn{1}{c}{PLCC$\downarrow$ / SRCC$\downarrow$} & \multicolumn{1}{c}{PLCC$\downarrow$ / SRCC$\downarrow$}\\
\midrule
Square Attack & & 0.635 / 0.579 & 0.617 / 0.564 & 0.570 / 0.521 \\
\midrule
AttackVQA   & & \textbf{0.335 / 0.289}	 & 0.429 / 0.384 & 0.479 / 0.392 \\
\midrule
&NIMA&  0.578 / 0.518 & 0.546 / 0.470 & 0.531 / 0.514 \\

PGD&PaQ-2-PiQ &0.619 / 0.571 & 0.586 / 0.341 & 0.598 / 0.516 \\

&SPAQ       &0.544 / 0.564 & 0.608 / 0.486 & 0.480 / 0.492 \\
\midrule
&NIMA&  0.475 / 0.453 & \textbf{0.369} / \underline{0.348} & \textbf{0.426} / \underline{0.419} \\

IC2VQA (ours) &PaQ-2-PiQ &0.450 / 0.404 & 0.414 / 0.396 & 0.459 / 0.428 \\

&SPAQ       &\underline{0.404 / 0.311} & \underline{0.390} / \textbf{0.299} & \underline{0.439} / \textbf{0.366} \\
\bottomrule
\end{tabular}
\begin{tablenotes}
    \small $^\ast$Image quality metric used in the proposed method. For the PGD --- metric which is attacked. For the IC2VQA --- component of cross-layer loss. In the IC2VQA attack, the image quality metric specified in the table was utilized in conjunction with CLIP and temporal losses.
\end{tablenotes}
\end{center}

\end{table*}


\begin{table}[h]
\centering
\caption{Comparison of variations of the IC2VQA attack with different configuration. The table presents the mean absolute values of PLCC and SROCC correlations across different epsilons between linearly decreasing vectors and attacked VQA scores. VSFA was used as VQA.}
\label{table losses}
\begin{tabular}{ccc}
\toprule
Loss & PLCC $\downarrow$ & SRCC $\downarrow$ \\
\midrule
$\mathcal{L}_{xlayer}$ & 0.849 & 0.800 \\
$\mathcal{L}_{xlayer}+\mathcal{L}_{CLIP}$ & \textbf{0.472} & \underline{0.430} \\
$\mathcal{L}_{xlayer}+\mathcal{L}_{CLIP}+\mathcal{L}_{temp}$ & \underline{0.515} & \textbf{0.354}  \\
\bottomrule
\end{tabular}

\end{table}

\begin{figure}[!h]
  \centering
   {\includegraphics[width=\linewidth]{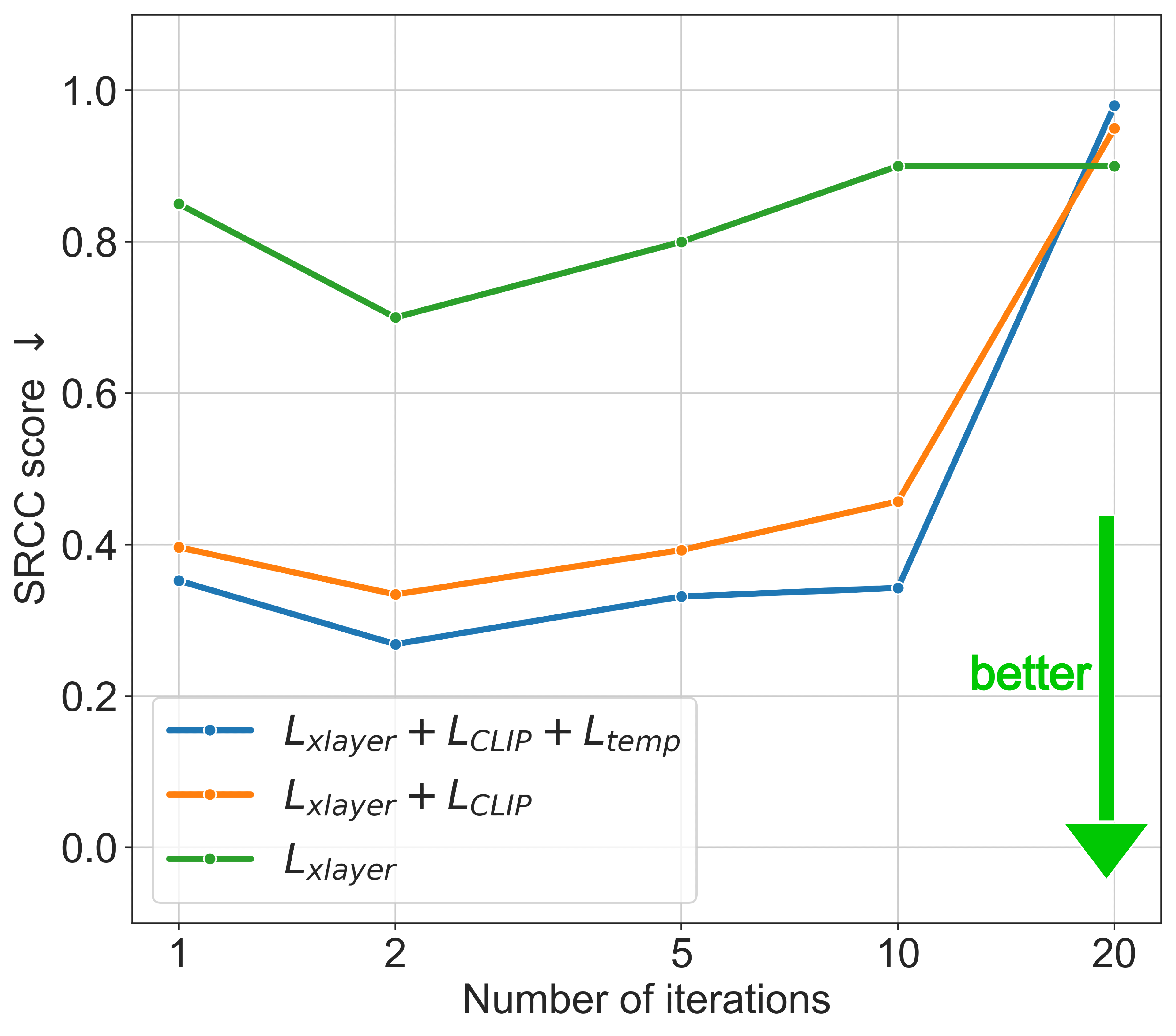}}
  \caption{The plot of variations of the IC2VQA attack under different configuration. The plot presents the median value of SRCC score across different epsilon with variation of the number of iterations.}
  \label{fig:graphic}
 \end{figure}

\begin{figure*}[!h]
  \centering
   {\includegraphics[width=\linewidth]{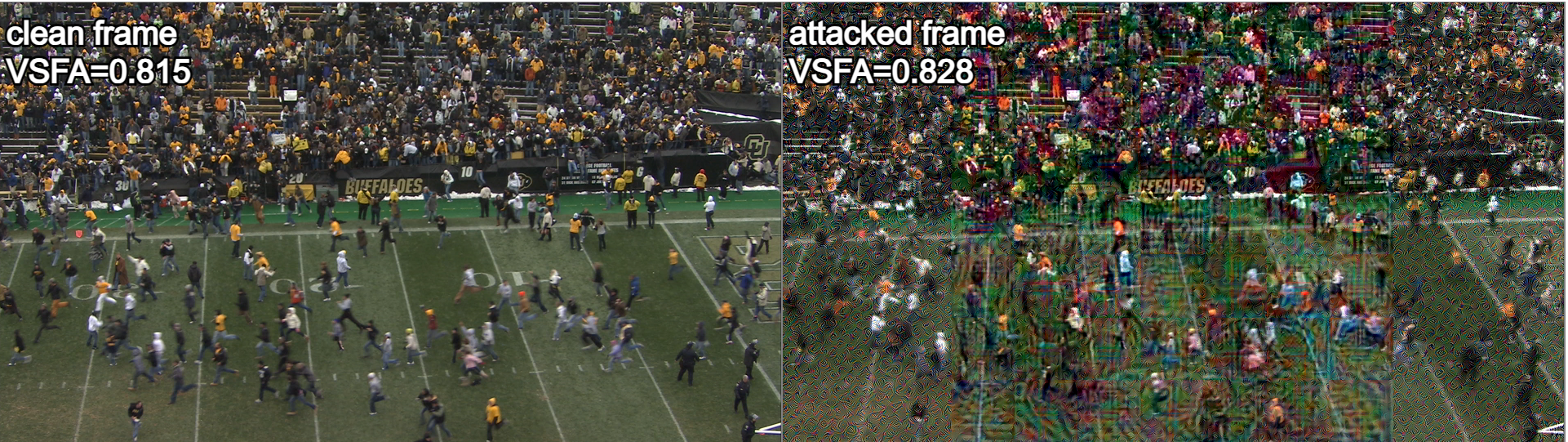}}
  \caption{Example of IC2VQA attack. Cross-layer loss is computed for layer1 of SPAQ, $\epsilon$ is set 50/255, number of iterations is set to 20. The visual quality of clean video is obviously higher than that of the attacked video, however, VSFA metric rates the attacked video as having higher quality.}
  \label{fig:example_png_attack}
 \end{figure*}

Results of comparison with other methods shown in the Table \ref{table comparison}. The proposed IC2VQA attack method demonstrated promising results across all three VQA models, achieving the reduction in PLCC and SRCC scores up to 0.425 and 0.380 on average, respectively. Additionally, it outperformed competing methods in two out of the three VQA black-box models. Furthermore, the results demonstrate that methods specifically designed for the VQA task, such as AttackVQA and the proposed IC2VQA, consistently outperform PGD and Square Attack, which are adaptations from classification tasks. This highlights the importance of developing approaches tailored for VQA challenges. Figure \ref{fig:example_png_attack} presents the example of the proposed attack. We can see that VQA metric fails to accurately assess the quality of the degraded video, assigning it a higher score.

 \begin{figure}[!h]
  \centering
   {\includegraphics[width=1.0\linewidth]{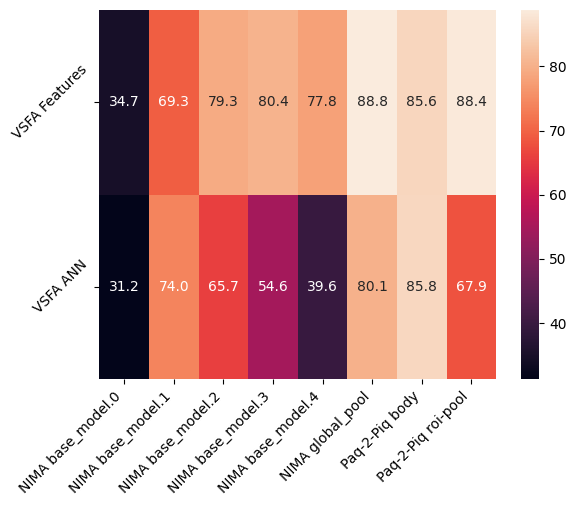}}
  \caption{Heatmap of cosine similarity between the features of VSFA layers and those from the NIMA and PaQ-2-PiQ layers. The values represent the cosine similarity scaled by a factor of 100.}
  \label{fig:heatmap_2}
 \end{figure}

\section{\uppercase{Ablation study}}
\subsection{Loss Configuration}
To experimentally demonstrate effectiveness of combination of losses in comparison with single $\mathcal{L}_{xlayer}$, we evaluated our attack in configuration with only one image quality metric $\mathcal{L}_{xlayer}$, with one image quality metric and CLIP image model $\mathcal{L}_{xlayer}+\mathcal{L}_{CLIP}$ and with one image quality metric, CLIP image model and temporal regularization $\mathcal{L}_{xlayer}+\mathcal{L}_{CLIP}+\mathcal{L}_{temp}$. In experiment we evaluate IC2VQA configurations on white-box models NIMA, PaQ-2-PiQ and SPAQ and black-box VSFA model and scored them by median absolute value of correlations. The results of the comparison are shown in the Table \ref{table losses} and Figure \ref{fig:graphic}. From Table \ref{table losses}, we observe that addition of cosine similarity between CLIP features ($\mathcal{L}_{CLIP}$) to the loss function enhances the attack's success by 1.8 times. The temporal loss increases the attack's success in terms of SRCC by 1.2 times and slightly decreases PLCC. Figure \ref{fig:graphic} shows that the combined loss function $\mathcal{L}_{xlayer}+\mathcal{L}_{CLIP}+\mathcal{L}_{temp}$ outperforms others in attack success, as measured by SRCC, across all iteration values.

The results of this experiment show that the addition of all components contributes to the effectiveness of the attack method. Therefore, in the final version of the attack, we use the $\mathcal{L}_{xlayer}+\mathcal{L}_{CLIP}+\mathcal{L}_{temp}$ loss function.

\subsection{Feature Correlation}

In this section, we analyze the correlations between features in the deep layers of IQA and VQA metrics. Figure \ref{fig:heatmap_2} presents the heatmap of correlations between features from the VSFA VQA model and the NIMA and PaQ-2-PiQ IQA models. We observe that these features are often highly correlated, highlighting the fact that addition of IQA modalities to black-box attack on VQA can boost transferability with a high likelihood of success.

\section{\uppercase{Conclusion}}
\label{sec:conclusion}
In this paper we propose the novel adversarial attack on VQA metrics that operates as a black-box. The proposed \methodname performs a cross-modal transferable attack that utilizes white-box IQA metrics and the CLIP model. The results of extensive experiments showed that \methodname generates adversarial perturbations that are more effective compared to previous approaches, significantly reducing the SRCC and PLCC scores of a black-box VQA model. The proposed method can serve as a tool for verifying VQA metrics robusthess to black-box attacks. Furthermore, the vulnerabilities identified in this study can contribute to the development of more robust and accurate VQA metrics in the future.

\section*{\uppercase{Acknowledgements}}
The research was carried out using the MSU-270 supercomputer of Lomonosov Moscow State University.


\newpage

\bibliographystyle{apalike}
{\small
\bibliography{example}}



\end{document}